\pdfoutput=1

\documentclass[11pt]{article}

\usepackage[final]{acl_latex}

\usepackage{pifont}
\usepackage{times}
\usepackage{latexsym}
\usepackage{amssymb}

\usepackage[T1]{fontenc}

\usepackage[utf8]{inputenc}

\usepackage{microtype}

\usepackage{inconsolata}

\usepackage{graphicx}

\usepackage{booktabs}
\usepackage{amsmath}

\title{\includegraphics[height=12pt]{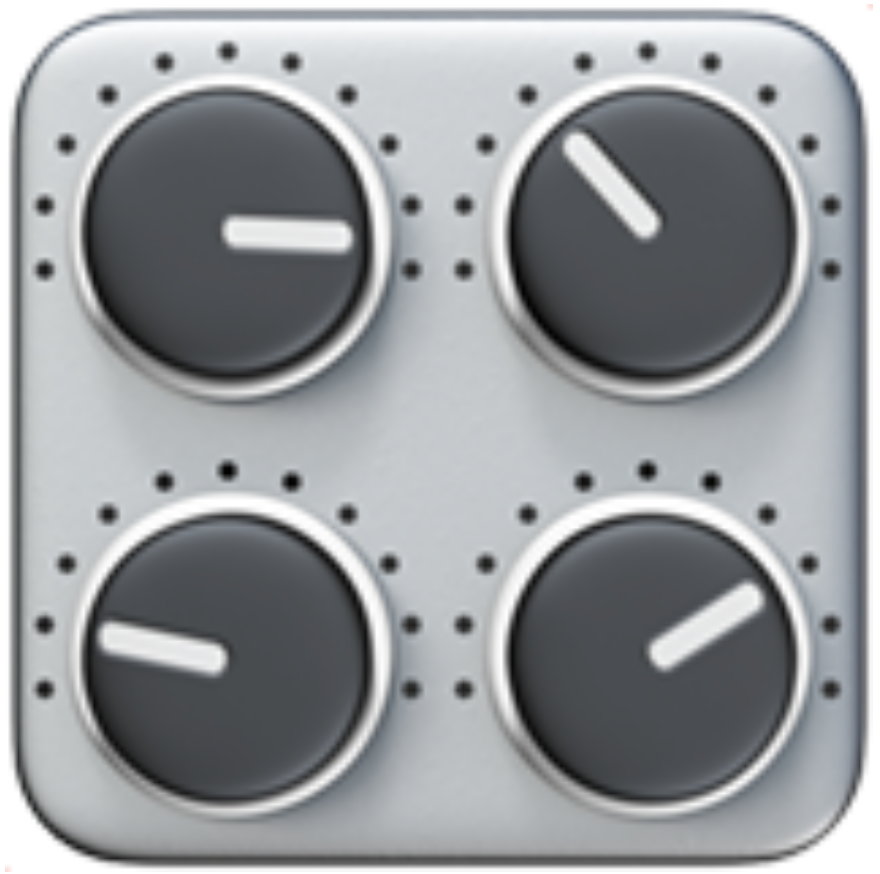} \texttt{EmoKnob}: Enhance Voice Cloning with Fine-Grained Emotion Control}

\author{
Haozhe Chen \\
  Columbia University \\
  \texttt{hc3295@columbia.edu} \\\And
  Run Chen \\
  Columbia University \\
  \texttt{runchen@cs.columbia.edu} \\\And
  Julia Hirschberg \\
  Columbia University \\
  \texttt{julia@cs.columbia.edu} 
  }

\begin{document}
\maketitle
\begin{abstract}
While recent advances in Text-to-Speech (TTS) technology produce natural and expressive speech, they lack the option for users to select emotion and control intensity. We propose \texttt{EmoKnob}, a framework that allows fine-grained emotion control in speech synthesis with few-shot demonstrative samples of arbitrary emotion. Our framework leverages the expressive speaker representation space made possible by recent advances in foundation voice cloning models. Based on the few-shot capability of our emotion control framework, we propose two methods to apply emotion control on emotions described by open-ended text, enabling an intuitive interface for controlling a diverse array of nuanced emotions. To facilitate a more systematic emotional speech synthesis field, we introduce a set of evaluation metrics designed to rigorously assess the faithfulness and recognizability of emotion control frameworks. Through objective and subjective evaluations, we show that our emotion control framework effectively embeds emotions into speech and surpasses emotion expressiveness of commercial TTS services.\footnote{See audio samples, code, and live demo at \href{https://emoknob.cs.columbia.edu}{\texttt{emoknob.cs.columbia.edu}}.}

\end{abstract}

\section{Introduction}
\begin{figure*}[t]
  \centering
  \includegraphics[width=\textwidth]{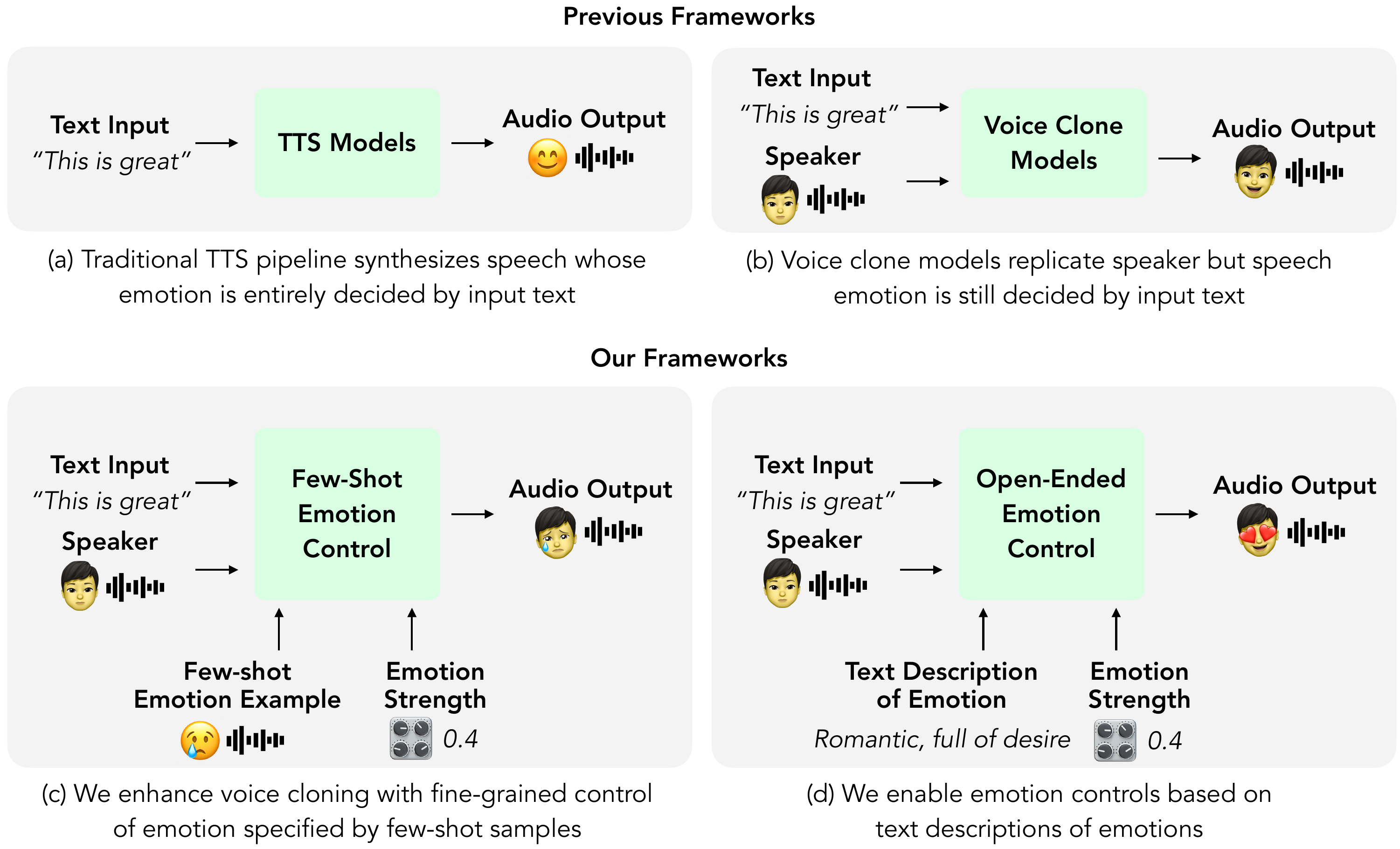}
  \caption{Fine-grained emotion control with \texttt{EmoKnob}. While existing TTS and voice cloning frameworks lack the option for users to control emotions in speech, our framework allows users to embed arbitrary emotion with a specified intensity in speech with few-shot samples. This framework allows us to propose two methods for controlling emotions based on open-ended text descriptions of emotions.}
  \label{fig:teaser}
\end{figure*}
The complexity of human communication extends far beyond mere verbal exchange. Vocal inflections and emotional undertones play pivotal roles in conveying meaning. While text alone can be ambiguous in meaning \citep{Jenkins2020Detecting}, different emotions in voices can articulate different messages in the same piece of text \citep{Nygaard2002Resolution}. Consider Shakespeare's iconic phrase, \textit{To be or not to be}. This line can express despair, contemplation, defiance, or resignation, depending on the speaker's emotional delivery, illustrating the profound impact of vocal emotions in communication.

The ultimate objective in the field of conversational systems is to develop intelligent agents capable of comprehending, deciding, and synthesizing speech with nuanced emotional undertones. While recent advances in Text-to-Speech (TTS) technology have achieved remarkable naturalness and expressiveness in synthesized voices\cite{elevenlabsTextSpeech, openaitts, microsoftTextSpeech}, these systems lack the capability for users to select and control the emotional tone and intensity. The emotion conveyed in the generated speech is solely determined by the text, without allowing for variability or intensity control.

Previous works on emotion control in speech synthesis primarily focus on a few simple emotion categories  \citep{DBLP:journals/corr/abs-2201-06460, LORENZOTRUEBA2018135, kang2023zetspeech, qin2024openvoice}. These methods do not allow control of a more diverse array of emotions. Synthesis for more complex and heterogeneous emotions like charisma \citep{charisma} and empathy \citep{chen2024detecting} is not well studied.

Our work leverages recent breakthroughs in foundation models for voice cloning \citep{MetaVoice, anastassiou2024seedtts, sunobark, casanova2024xtts, shen2018natural}. By exploring the rich expressiveness in these models' latent embedding spaces, we develop methods to extract a representation for any emotion with just a few demonstrative samples. These representations are inherently synergistic with the speech generation capabilities of rapidly advancing voice cloning/TTS models, enabling us to generate high quality speech while applying fine-grained emotion controls. This approach proves effective for both simple and complex emotions and includes mechanisms to adjust emotional intensity with a scalar knob.

Our framework's capability of applying fine-grained emotion control for any emotion with a few demonstrative examples enables us to propose two methods for applying emotion control based on arbitrary text descriptions of emotions. We use a synthetic-data-based and a retrieval-based method to leverage recent advances in Large Language Models (LLMs) and text embedding models \cite{chatgpt, salesforceairesearchSFREmbeddingMistralEnhance}, in conjunction with our few-shot emotion control framework, to address a lack of open-ended captioned emotional speech dataset.

We recognize that emotion control in speech synthesis is still at its early stage, and traditional evaluation metrics for TTS systems cannot comprehensively evaluate emotion control frameworks. We therefore introduce a set of rigorous evaluation metrics designed to systematically measure the effectiveness of an emotion control framework at faithfully conveying recognizable emotions. 

With a set of subjective and objective evaluations, we show that our framework produces faithful and recognizable emotion control on speech. We find that 83\% of the participants consider that speech with emotion enhancement by our framework surpasses leading commercial TTS services at conveying these emotions. 

\section{Related Work}

\begin{table}[h]
\centering\scriptsize
\begin{tabular}{@{}p{1.4cm} p{0.8cm} p{0.8cm} p{0.9cm} p{1.6cm}@{}}
\toprule
 & \begin{tabular}[c]{@{}l@{}}Expressive \\Emotion \\ Control\end{tabular} & \begin{tabular}[c]{@{}l@{}}Few-Shot \\Emotion\\ Control\end{tabular} & \begin{tabular}[c]{@{}l@{}}Open-Ended \\Emotion \\ Control\end{tabular} & \begin{tabular}[c]{@{}l@{}}Synergetic with\\ TTS Model \\Advances\end{tabular} \\ \midrule
\begin{tabular}[c]{@{}l@{}}Classifier-Based\\ Style Transfer$^1$ \end{tabular} & \checkmark & \ding{55} & \ding{55} & \ding{55} \\\hline
\begin{tabular}[c]{@{}l@{}}Domain \\Adversarial\\ Training $^2$\end{tabular} & \checkmark & \checkmark & \ding{55} & \ding{55} \\
\hline
\begin{tabular}[c]{@{}l@{}}Voice Text \\Descriptions$^3$\end{tabular} & \ding{55} & \ding{55} & \checkmark & \ding{55} \\ \hline
Ours & \checkmark & \checkmark & \checkmark & \checkmark \\ \bottomrule
\end{tabular}
\caption{Comparison between our framework and prior works on emotion control in speech synthesis. Our framework allows few-shot emotion control of arbitrary emotions and is synergetic with rapidly advancing text-to-speech models. We also propose two frameworks that allow users to control emotions with open-ended text emotion description. \scriptsize{$^1$\citet{DBLP:journals/corr/abs-2201-06460, LORENZOTRUEBA2018135, kang2023zetspeech, qin2024openvoice}. $^2$ \citet{10095619}. $^3$\citet{guo2022prompttts, yang2023instructtts, lacombe-etal-2024-parler-tts, lyth2024natural}.}}
\end{table}

\subsection{Foundational Model for TTS and Voice Cloning}
\label{sec:related-work-foundational}
Large foundational models have become the basis of many machine learning fields such as text \citep{openai2024gpt4} and images \citep{radford2021learning}. These large foundational models are trained in an unsupervised manner with massive datasets and learn high quality representations of data, which are commonly used directly or through fine-tuning for downstream tasks. 

The TTS domain also sees a rising trend in large, foundational models. These end-to-end models trained on large corpora provide natural speech rendering from text. \citet{MetaVoice} trains a 1.2B parameter model with 100K hours of speech for TTS; \citet{Lajszczak2024} trains a 1B parameter model on 100K open-domain speech data. Many of these models are capable of replicating a speaker's voice in zero-shot or few-shots \citep{MetaVoice, anastassiou2024seedtts, sunobark, casanova2024xtts, shen2018natural}. Our work explores how to leverage the high quality speaker representation learned by these foundational models to enhance voice cloning with few-shot fine-grained emotion control. In particular, we focus on manipulating the latent speaker embedding in \citet{MetaVoice}.

\subsection{Emotion and Style Control in Speech Synthesis}
While models discussed in Section \ref{sec:related-work-foundational} and existing commercial services \citep{openaitts, microsoftTextSpeech, elevenlabsTextSpeech} produce natural sounding speech, their speech output's emotions are primarily decided by input text, and the emotion strength cannot be controlled. Users thus cannot select arbitrary emotions for a piece of text. However, emotions expressed through acoustic-prosody serve an important additional channel for conveying information \citep{Gobl2003The, Patel2011Mapping, Laukkanen1997On}. 

Previous work trains latent speech style space on small corpora and cannot generalize to style transfer beyond the training corpus \citep{zhang2019learning}. In addition, existing labeled emotional speech datasets \citep{osti_10250021, poria2019meld} are limited to a few categories of basic emotions. 
Previous work thus commonly bases emotion controls on categorical emotion label inputs and is limited in types of emotions that can be controlled \citep{DBLP:journals/corr/abs-2201-06460, LORENZOTRUEBA2018135, kang2023zetspeech, qin2024openvoice}. Extending these methods to control new emotions require extensive retraining of models, preventing expressive emotion control over many emotions. These methods' requirement on large labeled datasets also prevents emotional control on more complex, nuanced emotions represented by more specialized, heterogeneous datasets such as charisma \citep{charisma} and empathy \citep{chen2024detecting}. 

While \citet{10095619} uses domain adversarial training to achieve few-shot emotion transfer, their method requires training a style encoder built from scratch and is not compatible with existing and future large foundational models. Thus, it is unable to improve naturalness and expressiveness in current and future TTS model developments. Our work provides a training-free framework that leverages a foundation model's TTS capability for single/few-shot emotion control and is inherently synergetic with growing foundation speech models.

\subsection{Open-ended Text Prompt Control on Voice}
A recent strand of works use text description to control voices. \citet{guo2022prompttts, yang2023instructtts, lacombe-etal-2024-parler-tts, lyth2024natural} allow users to describe qualities such as tone, pitch, gender, and emotions of a voice before synthesizing speech with the described voice. While existing speech datasets lack text descriptions of voices, this work bypasses this obstacle by creating synthetic text captions based on acoustic-prosodic features and speaker metadata. These methods do not generalize well to text descriptions beyond the format and the scope of the synthetic captions. The emotion control with these methods is limited to the categorical emotion labels in speaker metadata. These methods also do not allow voice cloning and emotion variation on an unseen speaker. Based on our method's capability to enhance voice cloning with single/few samples, we propose retrieval and synthetic data based frameworks for synthesizing expressive emotions with open-ended text descriptions.

\section{Methods}
\begin{figure*}[t]
  \includegraphics[width=\textwidth]{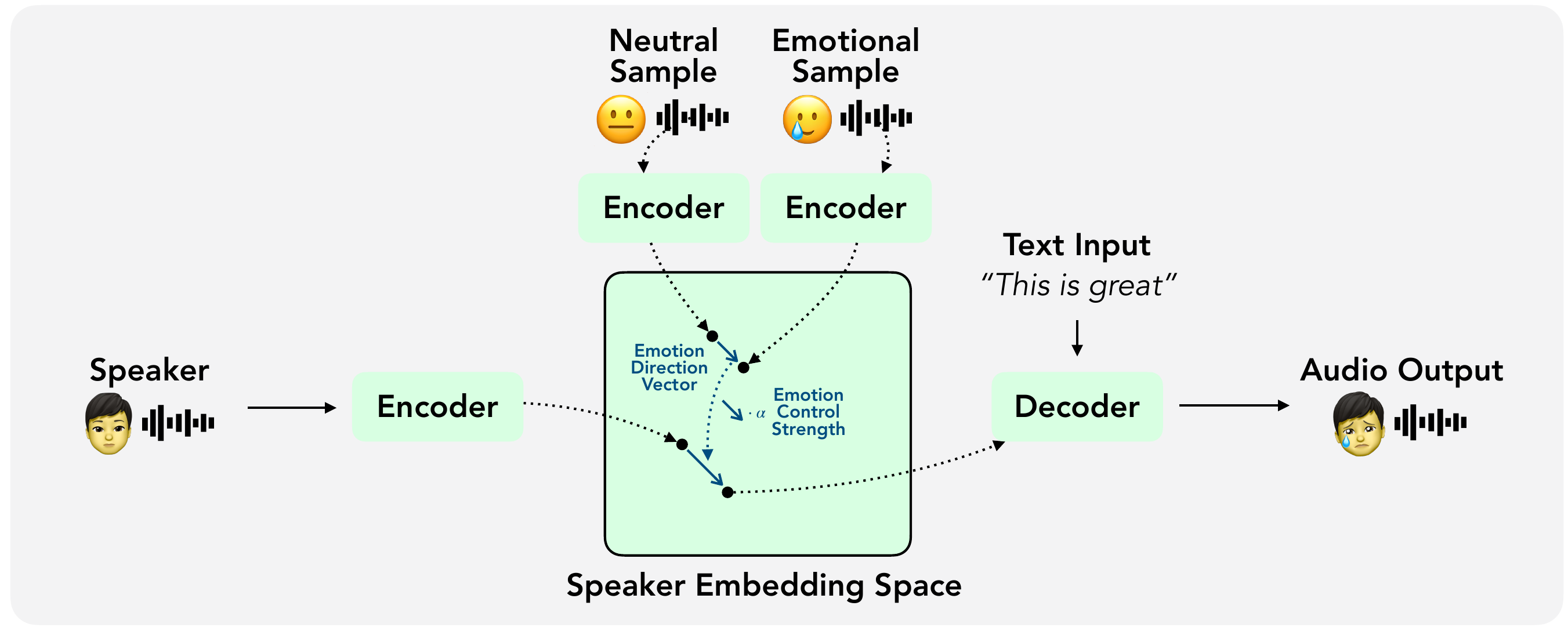}
  \caption{\texttt{EmoKnob}'s few-shot emotion control pipeline. \texttt{EmoKnob} first extracts an emotion direction vector in speaker embedding space of pre-trained foundation voice cloning models with a pair of neutral and emotional sample. Then, \texttt{EmoKnob} manipulates the reference speaker's embedding with the obtained emotion direction vector and a specified emotion strength to embed the emotion into speech.}
  \label{fig:method-few-shot}
  \vspace{-7mm}
\end{figure*}

We apply fine-grained emotion controls by manipulating the speaker embedding space of pre-trained foundation voice cloning models. This framework allows us to apply emotion control with few-shot emotional speech samples. The few-shot capability enables us to design two frameworks for applying control with emotion specified by arbitrary text descriptions.

\subsection{Preliminaries: Pre-Trained Foundational Voice Cloning Model}
Existing voice cloning models \citep{MetaVoice, anastassiou2024seedtts, sunobark, casanova2024xtts, shen2018natural} can be abstracted into a two-stage architecture with a speaker encoder $E$ that takes in a speaker reference clip $x_s$ and outputs a speaker embedding $u_s$. A conditional text-to-speech decoder $D$ then takes in input text $I$ to output speech audio $y_{s,I} = D(u_s, I)$ that utters $I$ replicating speaker's voice. We will manipulate the speaker embedding space (output space of $E$ and conditional input space of $D$) to obtain an emotion representation and obtain few-shot emotion control.

\vspace{-3mm}
\subsection{Few-Shot Fine-Grained Emotion Control}
\vspace{-2mm}
We hypothesize that a pre-trained foundation voice cloning model's speaker embedding provides expressive representations for acoustic-prosodic qualities. Our framework disentangles how an speaker embedding represents speaker-specific qualities and speaker-independent emotions. We then use the  speaker-independent emotions obtained to apply fine-grained emotion control on an arbitrary speaker representation. We show this process for few-shot fine-grained emotion control in Figure \ref{fig:method-few-shot}.

We disentangle speaker-specific qualities and speaker-independent emotion representations by using paired samples of emotional speech $x_e^i$ and neutral speech $x_n^i$ from the same speaker. We encode representations $u_e^i, u_n^i$ for these $i$-th pairs of samples in a speaker embedding space with the pre-trained speaker encoder $E$: $u_e^i = E(x_e^i), u_n^i = E(x_n^i)$.

We hypothesize that taking their difference results in a speaker-independent emotion direction vector $v_e^i$. In addition, we normalize $v_e^i$ for convenient fine-grained emotion strength control later: $v_e^i = \frac{u_e^i - u_n^i}{||u_e^i - u_n^i||}$

We can obtain the emotion direction vector by averaging over many pairs of samples. We will show in experiments that single-shot ($N=1$) suffices to produce high-quality emotion control in many cases:
\[ v_e = \frac{1}{N} \sum_{i=1}^{N} \frac{u_e^i - u_n^i}{||u_e^i - u_n^i||}  \]

Given a new speaker reference sample $x_s$, we hope to replicate the speaker's voice qualities while controlling emotions in an utterance. We first obtain the reference speaker's speaker embedding with $u_s = E(x_s)$. Then, we apply emotion control with
\[ u_{s,e} = u_s + \alpha \cdot v_e \]
where emotion control strength $\alpha$ is a scalar that enables fine-grained control of emotion intensity. We hypothesize that larger $\alpha$ values lead to more intense emotions in the speech produced.

Finally, we use pre-trained decoder $D$ to synthesize $y_{s,I,e}$, a speech utterance of text $I$ replicating speaker $s$'s voice while conveying emotion $e$.

\subsection{Towards Open-Ended Text Prompted Emotion Control}
\begin{figure*}[t]
    \hspace{-13mm}
  \includegraphics[width=1.15\textwidth]{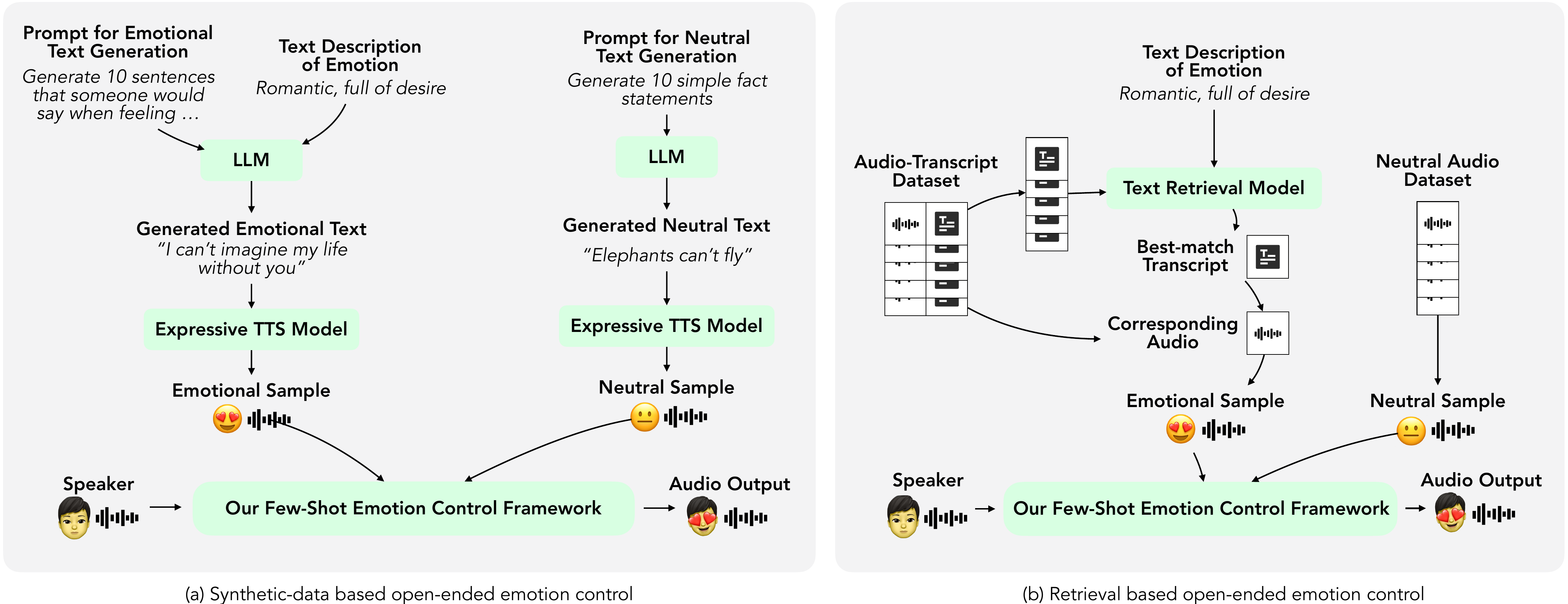}
  \caption{\texttt{EmoKnob} enables emotion control with open-ended text descriptions of emotion. Based on recent advances in LLMs and \texttt{EmoKnob}'s capability of applying emotion control with few-shot samples, we propose two methods that bypass the data insuffiency problem in emotional speech and embed emotions described by open-ended text descriptions into speech.}
  \label{fig:method-open-ended}
  \vspace{-5mm}
\end{figure*}

Our framework's ability to apply emotion controls with the few-shot demonstration allows us to design two frameworks that take in open-ended text description of an emotion and apply fine-grained control on output speech for the specified emotion. These frameworks allow synthesis of speech with emotions such as \textit{Romantic, full of desire} and \textit{Grateful, appreciative, thankful, indebted, blessed} that are nuanced in details and lack existing datasets. Both frameworks take advantage of recent development in LLMs to overcome the lack of a labeled emotional speech dataset.

\subsubsection{Synthetic Data-Based Method}
While existing TTS models and services do not allow emotion control, they produce expressive and accurate emotions for texts that obviously convey the emotions \citep{openaitts, microsoftTextSpeech, elevenlabsTextSpeech}. We leverage this quality to generate synthetic emotional samples that can be used for emotion control with our framework. We show this process in Figure \ref{fig:method-open-ended}(a).

Given a text description $T$ of an emotion $e$, we prompt an LLM to generate $N$ text samples $I^{1,\cdots,N}_e$, that obviously convey the emotion: $ I^{1,\cdots,N}_e = \textrm{LLM}(T)$.
Prompted with prompts such as \textit{Generate 10 sentences that someone would say when feeling [emotion]}, LLM generates emotional texts that conveys emotion $e$. Then, we use expressive commercial TTS services to obtain an emotional speech sample $x_e^{1\cdots i}$: $ x_e^i = \textrm{TTS}(x_e^i) $.

We can obtain neutral audio samples with the same procedure by first prompting LLM with prompts such as \textit{Generate 10 simple fact statements} to generate neutral texts. Then, we can obtain the neutral audio samples $x_n^i$ with the TTS services.

We can then use the  emotional speech samples obtained $x_e^i$ and $x_n^i$ with our few-shot emotion control framework to apply fine-grained emotion control on new speakers.


\subsubsection{Transcript Retrieval-Based Method}
While a synthetic-data-based method enables open-ended emotion control while bypassing the lack of captioned emotion datasets, the high cost of expressive TTS services limits the framework's wide usage. In this section, we hypothesize that in existing datasets with speech-transcript pairs, transcripts that obviously convey an emotion are matched with audio clips that convey the emotion. We leverage recent developments of text embedding models and document retrieval pipeline to find emotional audio samples that we can use for few-shot emotion control. We show this pipeline in Figure \ref{fig:method-open-ended}(b).

Given a text description $T$ of an emotion $e$ and a text embedding model $M$, we retrieve transcript-audio pairs $(I_e^j, x_e^j)$ in a dataset such that the transcript $I_e^j$ best matches the emotion description: $ j = \arg\max_j M(I_e^j)^T M(T) $.

We can find neutral samples $x_n$ either with the same retrieval pipeline or neutral labels in the dataset, which are more widely available than diverse emotion labels.


\section{Experiments}

\subsection{Evaluation Metrics}

\subsubsection{Subjective Evaluations}
\label{sec:subjective-eval-desc}

Given the novelty of fine-grained emotion control in text-to-speech synthesis, there is not an established paradigm for examining this capability. To rigorously test the objective of providing fine-grained, faithful emotion control, we proposed the following subjective evaluation metrics:\\\\
\textbf{Emotion Selection Accuracy (ESA)}:  Participants compare audio samples with and without control generated from emotion-neutral text, selecting which better conveys the emotion. ESA measures the percentage choosing the controlled audio and tests the system's ability to embed any emotions to any text.\\
\textbf{Emotion Enhancement Accuracy (EEA)}: Participants compare audio samples with and without control generated from emotion-matching text, selecting which better conveys the emotion. EEA measures the percentage choosing the controlled audio and tests the method's ability to amplify text's emotions.\\
\textbf{Emotion Discrimination Test (EDT)}: Participants compare two audio samples generated from the same neutral text and controlled with different emotions, selecting the one matching a given emotion. EDT evaluates the distinguishability and faithfulness of emotion control. \\
\textbf{Emotion Identification Test (EIT)}: Participants identify the emotion in a controlled audio sample from neutral text, choosing between two emotion labels. EIT measures the accuracy of emotion identification and verifies the recognizability of emotions resulted from emotion control.\\
\textbf{Emotion Selection Comparison (ESC)}: Participants compare our emotion-controlled audio to commercial TTS audio with neutral text, selecting which conveys more specified emotion. ESC measures percentage of selecting our controlled audio and evaluates system advantage over existing TTS services to embed any emotion into any text.\\
\textbf{Emotion Enhancement Comparison (EEC)}: Similar to ESC, but with emotion-matching text. EEC evaluates emotion expressiveness after control compared to commercial TTS without emotion control functionality. \\
\textbf{Emotion Strength Test (EST)}: Participants compare two audio samples controlled with the same emotion but different emotion strengths $\alpha$, selecting which conveys more emotion. EST measures correct response percentage and evaluates our framework's effectiveness at fine-grained control over emotion intensity.

Since all metrics are calculated from binary choice questions, 50\% serves as the random guess baseline to all metrics. We asked one question for each emotion and each metric to 23 university student volunteers from our lab and recruited on campus. Participants are told that responses are used to evaluate a new emotional text-to-speech framework. This study is approved by IRB. We anonymized the participant response. We provided the full subjective evaluation survey we used at \url{https://frolicking-baklava-af4770.netlify.app/}. For EEC and ESC, we compared speech generated from our framework with speech generated with ElevenLabs \citep{elevenlabsTextSpeech}.

\subsubsection{Objective Evaluation}
Since our goal is to preserve source speaker identity and maintain accurate text-to-speech synthesis while conducting emotion control, we follow previous voice cloning work \citep{anastassiou2024seedtts, shah2023parrottts} on measuring word error rate (WER) and speaker similarity (SIM). We use 100 texts from Common Voice dataset \citep{ardila2020common} to calculate WER and SIM. 

For WER, we first transcribe the generated clips with Whisper-large-v3 \citep{radford2022robust} and calculate WER with jiwer library \citep{githubGitHubJitsijiwer}. We use the WER of audio generated without any emotion control (original voice cloning model) as a baseline of comparison. Similar WER between emotion-controlled audio and baseline suggests that our framework preserves the high quality TTS in base voice cloning models. 

For SIM, we used spkrec-ecapa-voxceleb \citep{speechbrain} to measure the similarity between generated audio and a reference speaker clip. We use SIM between audio generated without any emotion control and a reference speaker clip as baseline. Similar SIM between the baseline and using emotion-controlled audio suggests our framework's faithful replication of reference speaker while applying emotion control.

\subsection{Experiment Details}

We use MetaVoice-1B \citep{MetaVoice} as the base voice cloning model, while our framework can be easily extended to any embedding-conditioned voice cloning model. We conduct speech generation on a single NVIDIA L40 GPU. We use an additional NVIDIA L40 GPU for text retrieval in text-retrieval based open-ended emotion control.
We provide an audio sample page at \href{https://emoknob.cs.columbia.edu}{\texttt{emoknob.cs.columbia.edu}}. 

\subsection{Single-Shot Control of Simple Emotions}

We first show our framework's effectiveness on fine-grained emotion control with simple emotion categories: \textit{Happy, Surprise, Angry, Sad, Disgust, Contempt}. We obtain an emotion direction vector in single-shot (one pair of same-speaker emotional and neutral speech clips) from the MSP Podcast dataset \citep{Lotfian_2019_3}. We fix emotion strength $\alpha$ to be 0.4 for all samples in evaluation.

We report the subjective evaluation results in Table \ref{tab:simple-subj-eval} and the objective evaluate results with a standard deviation in Table \ref{tab:simple-obj-eval}. High ESA and ESC values shows that our emotion control framework is capable of embedding arbitrary emotion in any text, surpassing commercial TTS services without emotion control option. High EEA and EEC values show that our framework enhances emotions into emotion-matching text, surpassing emotion utterances of commercial TTS services. High EDT and EIT values show that our framework produces recognizable emotions in speech. High EST values show that the emotion strength $\alpha$ option in our framework faithfully produces different strengths of emotions specified by corresponding values.  

Speech produced from emotion control shows similar WER within uncertainty as a baseline of no emotion control and thus preserves high quality TTS of the base model. Emotion-controlled speech also shows similar SIM within uncertainty as the baseline, showing that our framework preserves speaker identity well while conducting emotion control.

\begin{table}[h!]
\centering
\scriptsize
\begin{tabular}{lp{0.3cm}p{0.3cm}p{0.3cm}p{0.3cm}p{0.3cm}p{0.3cm}p{0.3cm}}
\toprule
 & ESA$\uparrow$ & EEA$\uparrow$ & EDT$\uparrow$ & EIT$\uparrow$ & ESC$\uparrow$ & EEC$\uparrow$ & EST$\uparrow$\\
\midrule
Happy   &  100\% & 100\% & 100\% & 100\% & 100\% & 100\% & 83\% \\
Surprise & 100\% & 100\% & 91\%  & 44\%  & 100\% & 61\%  & 91\% \\
Angry     &82\%  & 100\% & 82\%  & 74\%  & 100\% & 100\% & 100\% \\
Sad       &100\% & 83\%  & 91\%  & 100\% & 100\% & 83\%  & 74\% \\
Disgust   &74\%  & 91\%  & 91\%  & 74\%  & 61\%  & 83\%  & 91\% \\
Contempt  &61\%  & 83\%  & 13\%  & 52\%  & 74\%  & 74\%  & 74\% \\
\midrule
Averages & 86\% & 93\% & 78\% & 74\% & 89\% & 83\% & 86\% \\
Baseline  & 50\% & 50\% & 50\% & 50\% & 50\% & 50\% & 50\% \\
\bottomrule
\end{tabular}
\caption{Subjective evaluation results for emotion controls with simple emotions.}
\label{tab:simple-subj-eval}
\vspace{-5mm}
\end{table}

\begin{table}[h!]

\centering\scriptsize
\begin{tabular}{lcc}
\toprule
 & WER $\downarrow$ & SIM $\uparrow$ \\
\midrule
Happy     & 0.143 $\pm$ 0.349& 0.662 $\pm$ 0.087 \\
Surprise  & 0.061 $\pm$ 0.107& 0.703 $\pm$ 0.076 \\
Angry     & 0.082 $\pm$ 0.211 & 0.712 $\pm$ 0.060\\
Sad       & 0.113  $\pm$ 0.297& 0.719 $\pm$ 0.059\\
Disgust   & 0.05 $\pm$ 0.139& 0.719 $\pm$ 0.063\\
Contempt  & 0.053 $\pm$ 0.098& 0.712 $\pm$ 0.069\\
Average   & 0.085 $\pm$ 0.208 & 0.705 $\pm$ 0.077\\
\hline
w/o Emotion Control & 0.079$\pm$ 0.160 & 0.719 $\pm$ 0.071\\
\bottomrule
\end{tabular}
\caption{Objective evaluation results for controls with simple emotions.}
\label{tab:simple-obj-eval}
\end{table}

\subsection{Two-Shot Control of Complex Emotion}
Our framework allows a few-shot transfer of emotion onto new speakers and bases such transfer on expressive representation of foundation voice cloning models. We show that these features enable previously not studied controls on more complex, composite, and nuanced emotions. Our experiments focus on two emotions with corresponding datasets: (1) \textit{charisma} defined as conveying the personality of leadership and persuasiveness \citep{charisma}; and (2) \textit{compassionate empathy} defined as understanding another’s pain as if we are having it ourselves and taking action to mitigate problems producing it \citep{chen2024detecting}. For each emotion, we use two pairs of emotional and neutral speech from two speakers. We fix emotion strength $\alpha=0.4$ for all samples. 


We report the subjective and objective evaluation results in \ref{tab:complex-eval}. Subjective evaluation results show that our framework produces recognizable, faithful emotion selection and enhancement, surpassing commercial TTS on uttering specified emotions. Speech produced from emotion control shows similar WER and SIM within uncertainty as the baseline of no emotion control, showing that our framework preserves accurate TTS of the base model and speaker identity while conducting emotion control. 

\begin{table}[h!]

\centering\scriptsize
\begin{tabular}{lp{0.4cm}p{0.4cm}p{0.4cm}p{0.4cm}p{1.4cm}p{1.4cm}}
\toprule
 & ESA$\uparrow$ & EEA$\uparrow$ &ESC$\uparrow$ & EEC$\uparrow$ & WER$\downarrow$ & SIM$\uparrow$ \\
\midrule
Empathy   & 74\% & 83\% & 100\% & 22\% & 0.074$\pm$ 0.077 & 0.712$\pm$ 0.063  \\
Charisma  & 83\% & 91\% & 74\% & 74\% & 0.031$\pm$ 0.088 & 0.680 $\pm$ 0.070\\
\hline
Baseline  & 50\% & 50\% & 50\%& 50\% & 0.079 $\pm$ 0.160& 0.719 $\pm$ 0.071\\
\bottomrule
\end{tabular}
\caption{Subjective and objective evaluation results for controls with complex emotions}
\label{tab:complex-eval}
\end{table}

\subsection{Synthetic Data-Based Open-Ended Emotion Control}

\begin{table}[h!]
\centering\scriptsize
\begin{tabular}{lp{0.5cm}p{0.5cm}p{0.5cm}p{0.5cm}p{1.4cm}p{1.4cm}}
\toprule
 & ESA$\uparrow$ & EEA$\uparrow$ &ESC$\uparrow$ & EEC$\uparrow$ & WER$\downarrow$ & SIM$\uparrow$ \\
\midrule
Desire    & 61\% & 61\% & 61\% & 83\% & 0.066$\pm$ 0.132 & 0.713 $\pm$ 0.075 \\
Envy      & 83\% & 74\% & 61\% & 74\% & 0.085$\pm$ 0.131 & 0.704$\pm$ 0.071 \\
Romance   & 61\% & 91\% & 52\% & 91\% & 0.076$\pm$ 0.125 & 0.713$\pm$ 0.066 \\
Sarcasm   & 61\% & 61\% & 74\% & 74\% & 0.120$\pm$ 0.200 & 0.717$\pm$ 0.079 \\
\hline
Baseline  & 50\% & 50\% & 50\%& 50\% &0.079 $\pm$ 0.160& 0.719$\pm$ 0.071 \\
\bottomrule
\end{tabular}
\caption{Subjective and objective evaluation results for open-ended controls with emotion text descriptions through a synthetic data-based method.}
\label{tab:eval-synthetic}
\end{table}

We experiment with our synthetic-data based framework for emotion control on arbitrary text emotion description with emotions that do not have previously collected labeled datasets for emotional speech synthesis: \textit{Desire, enviousness, romance, sarcasm}. We use GPT4-o \citep{chatgpt} to generate emotional and neutral speech texts. We use ElevenLabs \citep{elevenlabsTextSpeech} to generate 10 pairs of samples (10 speakers) for each emotion. We fix emotion strength of $\alpha = 0.4$ for all samples. 

We report the subjective and objective evaluation results in Table \ref{tab:eval-synthetic}. Subjective evaluations indicate our recognizable and faithful emotion control in speech, outperforming commercial TTS in expressing specific emotions. Additionally, speech from our emotion control maintains similar WER and SIM to the baseline, confirming that our framework effectively preserves the base model's accuracy and speaker identity while controlling emotions.

\subsection{Retrieval-Based Open-Ended Emotion Control}
Since a text retrieval model works  best with descriptive, detailed texts, we focus on longer emotion descriptions of three emotions that lack established labeled datasets shown in Table \ref{tab:eval-retrieval}. We prefix the emotion descriptions with the retrieval prompt of \textit{Given a description, retrieve relevant transcript lines whose overall style/emotions matches the description} to enable retrieval models focused on the overall emotion of the transcript and avoid keyword matching. We use SFR-Embedding-Mistral \citep{salesforceairesearchSFREmbeddingMistralEnhance} as the text embedding model. We use 10 pairs of emotional and neutral samples for each emotion. We fix emotion strength $\alpha=0.5$ for all samples.

We report the subjective evaluation results and objective evaluate results with a standard deviation in Table \ref{tab:eval-retrieval}. The evaluation results show that our framework produces recognizable, faithful emotion selection and enhancement while preserving base model accuracy and reference speaker identity.
\begin{table}[h]
\centering\scriptsize
\hspace{-8mm}
\begin{tabular}{lp{0.2cm}p{0.4cm}p{0.4cm}p{0.4cm}p{1.4cm}p{1.4cm}}
\toprule
 & ESA$\uparrow$ & EEA$\uparrow$ &ESC$\uparrow$ & EEC$\uparrow$ & WER$\downarrow$ & SIM$\uparrow$ \\
\midrule
Grateful \textsuperscript{...} $^1$ & 83\% & 83\% & 83\% & 61\% & 0.146$\pm$ 0.387 & 0.650$\pm$ 0.072 \\
Curious, \textsuperscript{...} $^2$ & 61\% & 100\% & 61\% & 22\% & 0.124$\pm$ 0.298 & 0.655$\pm$ 0.060 \\
Blaming & 65\% & 69\% & 74\% & 74\% & 0.112$\pm$ 0.211 & 0.630$\pm$ 0.062 \\
Desire \textsuperscript{...} $^3$  & 78\% & 100\% & 100\% & 91\% & 0.062$\pm$ 0.144 & 0.664$\pm$ 0.091 \\
\hline
Baseline  & 50\% & 50\% & 50\%& 50\% & 0.079 $\pm$ 0.160& 0.719$\pm$ 0.071\\
\bottomrule
\end{tabular}
\caption{Subjective and objective evaluation results for open-ended controls with emotion text descriptions through retrieval-based methods. \scriptsize{}{$^1$ Grateful, appreciative, thankful, indebted, blessed. $^2$ Curious intrigued. $^2$ Desire and excitement.}}
\label{tab:eval-retrieval}
\vspace{-5mm}
\end{table}
\section{Ablation Studies}
\begin{figure}[t]
  \includegraphics[width=\columnwidth]{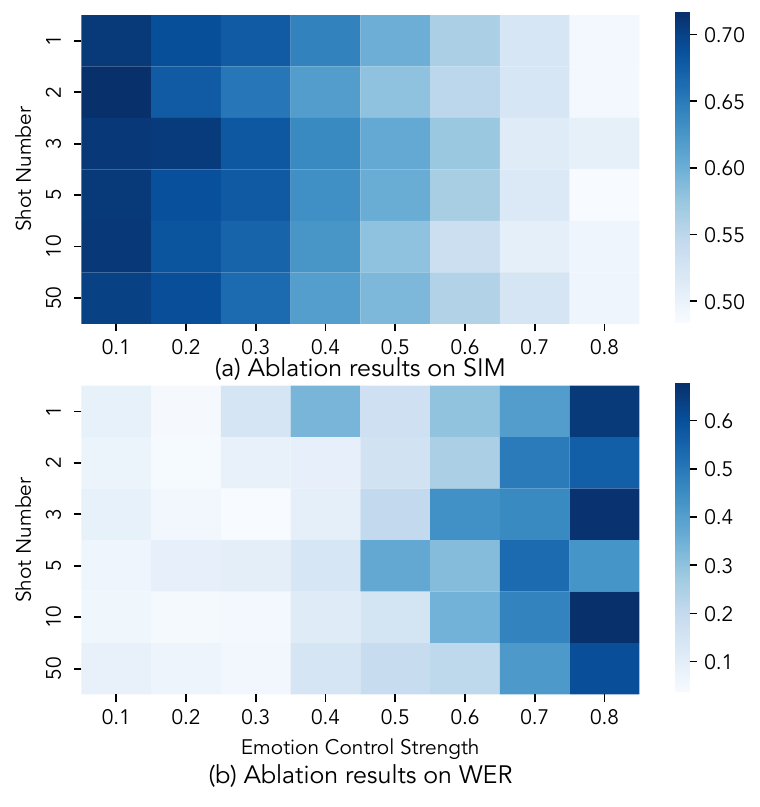}
  \caption{Ablation results measuring SIM and WER with varying shot number and emotion strength.}
  \label{fig:ablation}
  \vspace{-6mm}
\end{figure}

In this section, we conduct ablation studies that vary the shot (sample) number when obtaining the emotion direction vector and the emotion control strength $\alpha$ when applying emotion control. These ablation studies help users decide how to select these hyper-parameters. We report in Table \ref{fig:ablation} SIM and WER with audio generated with 100 Common Voice texts while applying emotion control of simple emotions with varying shot numbers and emotion strengths. We observe that both SIM and WER are insensitive to shot number and degrades as emotion control strength increases. Users thus need to trade off between generating more emotional clip with higher emotion strength and accurate TTS and voice clone. However, a larger number of samples make the method more robust in larger emotion control strength. Users thus could employ a larger number of samples to compensate the TTS quality decrease while obtaining more emotional speech.


\section{Conclusion and Future Works}
We proposed \texttt{EmoKnob}, a framework that enables fine-grained emotion control in voice cloning with few-shot samples. We also propose a synthetic-data-based and a retrieval-based method to embed emotions described by open-ended text into speech synthesis. Given novelty of the emotion control domain, we proposed a set of metrics to rigorously evaluate faithfulness and recognizability of emotion control. Our method establishes a new way of extracting emotion representation in foundation speech models thus bypassing data limitations. Future works can further explore emotion control paradigms such as synthesizing emotions in conversation turns based on these representations.



\section*{Limitations}
Naturalness and expressiveness of speech created by our framework is constrained by base voice cloning model. However, since we are seeing rapid advances in foundation speech models, and our method is inherently synergetic with these advances, speech produced by \texttt{EmoKnob} will naturally improve as voice cloning models scale up and improve. 

\section*{Potential Risks}
Risks in speech identity theft in voice cloning apply to our work. Practices such as voice cloning detection \citep{Malik2019Securing} and phasing out voice-based authentication systems \citep{openai-voice-challenges} help mitigate risks of our works.

\section*{Acknowledgements}
This research is being developed with funding from the Columbia Center of AI Technology (CAIT) research award and the Defense Advanced Research Projects Agency (DARPA). The views, opinions and/or findings expressed are those of the author and should not be interpreted as representing the official views or policies of the Department of Defense or the U.S. Government.
\bibliography{acl_latex}

\appendix



\end{document}